# Individual corpora predict fast memory retrieval during reading

**Markus J. Hofmann, Lara Müller, Andre Rölke, Ralph Radach**
Bergische Universität Wuppertal, Max-Horkheimer-Str. 20, 42119 Wuppertal
{mhofmann,lara.mueller,roelke,radach}@uni-wuppertal.de

**Chris Biemann**
Universität Hamburg, Vogt-Kölln-Straße 30, 22527 Hamburg
biemann@informatik.uni-hamburg.de

## Abstract

The corpus, from which a predictive language model is trained, can be considered the experience of a semantic system. We recorded everyday reading of two participants for two months on a tablet, generating individual corpus samples of 300/500K tokens. Then we trained word2vec models from individual corpora and a 70 million-sentence newspaper corpus to obtain individual and norm-based long-term memory structure. To test whether individual corpora can make better predictions for a cognitive task of long-term memory retrieval, we generated stimulus materials consisting of 134 sentences with uncorrelated individual and norm-based word probabilities. For the subsequent eye tracking study 1-2 months later, our regression analyses revealed that individual, but not norm-corpus-based word probabilities can account for first-fixation duration and first-pass gaze duration. Word length additionally affected gaze duration and total viewing duration. The results suggest that corpora representative for an individual's long-term memory structure can better explain reading performance than a norm corpus, and that recently acquired information is lexically accessed rapidly.

## 1 Introduction

There are three basic *stages* of memory (e.g. Paller and Wagner, 2002). All memories start with *experience*, which is reflected by text corpora (e.g. Hofmann et al., 2018). The training of a language model then reflects the process of memory *consolidation*. The final stage is memory *retrieval*, which can be examined in psycholinguistic experiments. In this paper, we measure the correlation of computational language modelling and cognitive performance.

We collected individual corpora from two participants reading on a tablet for two months and compared them to an extensive corpus mainly consisting of online newspapers (Goldhahn et al., 2012). To consolidate differential knowledge structures in long-term memory, word2vec models were trained from these corpora. For stimulus selection, we relied on these three language models to compute word probabilities and sentence perplexity scores for 45K sentences of a Wikipedia dump. Perplexity rank differences were used to select sentences with uncorrelated word probabilities for the three language models, allowing to estimate the independent contribution of the word probabilities in multiple regression analyses. The resulting 134 stimulus sentences were read by the participants in an eye tracking experiment. In the multiple regression analyses, we used these predictors to account for the durations of the first fixation on the words. We also predicted gaze durations, in which the duration of further fixations during first-pass reading are added. When the eye revisits a word after first-pass reading has been finished, the durations of further fixations are added into the total viewing duration (see Figure 1 for an overview of the present study). The underlying hypotheses of our research are that semantic expectancy has a top-down effect on word saliency at the visual level (Hofmann et al., 2011; Reilly & Radach, 2006), and words appearing in more salient contexts, are processed quicker by human subjects. Therefore, language models on individual reading corpora, realized e.g. by word2vec, should predict the processing speed.

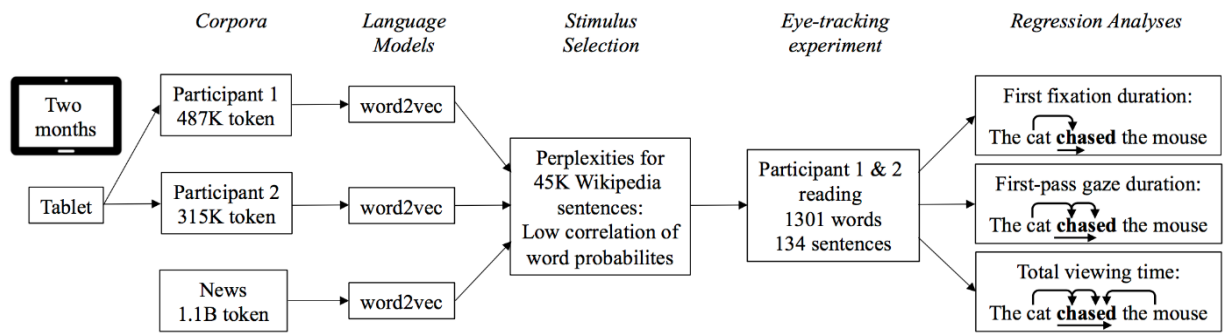

Figure 1. Overview of the present study

## 2 Learning history and memory consolidation

### 2.1 Norm corpora as a representative sample of human experience?

When selecting a corpus as a sample of the learning experience of human participants or language models, the question arises which corpus is most representative for which person. For instance, the knowledge of young adults is better characterized by corpora consisting of books written for younger adults, while older adults are more experienced – when therefore searching for corpora that account for their performance best, more diverse fictional and literature books are chosen (Johns, Jones, & Mewhort, 2018). Rapp (2014) proposed that corpus representativeness should be measured by the Pearson correlation of corpus-derived computational measures and an external measure of human performance. The knowledge of an average reader may be well represented by balanced corpora containing all sorts of content, such as Wikipedia. However, a previous study revealed that a newspaper corpus often provides higher correlations than Wikipedia when accounting for human cloze completion probabilities, as well as eye tracking or brain-electric data (Goldhahn et al., 2012; Hofmann et al., 2017). This result pattern could on the one hand be explained by the fact that typical German readers very frequently visit online newspapers. On the other hand, the encyclopedic nature of Wikipedia with one article per topic does not reflect the frequencies of exposure. Therefore, a newspaper corpus may be more representative. In his seminal work, Rapp (2014) avoided the possibility to observe the average language input of test persons, noting that it would be effortful and unpractical to collect corpora from the language of participants. In the present work, we directly address this problem by collecting individual corpora from two test participants, which is made possible by technological advances and cheaper hardware, i.e. tablets with cameras for eye-tracking.

When choosing a corpus, a key issue is its size. In general, a larger corpus may provide better word frequency estimates that allow for better human performance predictions than a small corpus (Rapp, 2014). For high frequency words, however, a corpus of 1 million words already allows for predictions comparable to larger corpora (Brysbaert & New, 2009). For low frequency words, performance predictions improve for corpora up to 16 million words, while there is hardly any gain for corpora greater 30 million words. Rather, extremely large corpora of more than a billion words may even decrease performance predictions. "For these sizes, it becomes more important to know where the words of the corpus came from" (Brysbaert & New, 2009). Movie subtitles represent spoken language, which humans typically encounter much more often than written language. In this case, the corpus is probably more representative, because it incorporates the more frequently produced and received language mode. Therefore, it appears that the representativeness for the language input can outperform sheer size.

Whether such smaller corpora are also sufficient to characterize contextual word probabilities has also been examined by Mandera et al. (2017). They showed that size does not always trump representativeness in predicting semantic priming, i.e. the facilitation of word recognition as a consequence of processing a preceding prime word. In general, the everyday social information represented in subtitles corpora may be more accessible for human subjects. Therefore, they tend to elicit effects in early eye-movement measures, particularly when combined with a language model that can generalize, such as neural network models, for instance (Hofmann et al., 2017, 2020).

Though personalized language signals are obviously used to optimize search engine performance, tempting us to purchase products by means of individualized advertisements, we are not aware of any scientific approach towards assessing corpus representativeness at the level of real individual participants. Jacobs (2019), however, showed that language models can well characterize fictional characters. He used SentiArt to analyze Harry Potter books and found that theory-guided contextual properties of the characters can provide a face valid approach to personality. Voldemort occurred in language contexts indicating emotional instability, thus he scored high in the pseudo-big-five personality trait of neuroticism. Harry Potter's personality, in contrast, can be most well characterized by the personality dimension of conscientiousness. Both, Harry and Hermione score high on the personality trait of openness to experience and intellect. With such a face valid characterization of the personality of fictional characters, we think that it is an ethical necessity to stimulate a scientific discussion about the potential of individual corpora, because they may tell us a lot about real persons.

## 2.2 Language models reflecting memory consolidation

Psycholinguistic reading and comprehension studies were dominated for a long time by latent semantic analysis (Deerwester et al., 1990). Pynte et al. (2008) showed that such a document-level approach to long-range semantics can better predict gaze duration than an earlier eye movement measure. In this case, however, the Dundee corpus was examined, in which discourse rather than single sentence reading was examined. Griffiths et al. (2007) showed that topics models may outperform LSA in psycholinguistic experiments, for instance by predicting gaze durations for ambiguous words (Blei et al., 2003). McDonald and Shillcock (2003), on the other hand, suggested that a word 2-gram model may reflect low-level contextual properties, given that they can most reliably account for first fixation duration rather than later eye-movement measures. Smith and Levy (2013) showed that a Kneser-Ney smoothed 3-gram model can also predict a later eye-movement measure, i.e. gaze duration, probably because a larger contextual window is used for the predictions of discourse comprehension using the Dundee eye-movement data set (cf. Pynte et al., 2008). Frank (2009) showed that a simple recurrent neural network is better suitable to address gaze duration data than a probabilistic context-free grammar (Demberg and Keller, 2008). The capability of neural networks to well capture syntax was also demonstrated by Frank and Bod (2011), who showed that an echo-state network better predicts gaze duration data than unlexicalized surprisal of particular phrase structures. For predicting first fixation durations of words that have been fixated only once during single sentence reading (Hofmann et al., 2017), a Kneser-Ney-smoothed 5-gram model provided good predictions, but a slightly better prediction was obtained by a recurrent neural network model (Mikolov, 2012). In the same work, Hofmann et al. (2017) showed that an LDA-based topic model (Blei et al., 2003) provided relatively poor predictions, probably because sentence- rather than document-level training more closely reflects the semantic short-range knowledge.

Since Bhatia (2017) and Mandera et al. (2017), word2vec models can be considered a standard tool for psycholinguistic studies. It is well known for eye-tracking research, that not only the predictability of the present, but also of the last and next word can influence fixation durations (e.g. Kliegl et al., 2006). Viewing durations of the present word can even be influenced to some extent by the word after the next word (Radach et al., 2013). As the present pilot study will be based on a limited number of observations, we used word2vec-based word embeddings trained to predict the probability of the present word by the default contextual window of the last and next two words (Mikolov et al., 2013). With such a contextual window of two during training, we intended to subsume the effects of the last and next words on the fixation duration of the present word during retrieval. Therefore, we decided to use this simple standard approach to natural language processing.

## 3 Memory retrieval in eye-tracking analyses

As has been summarized in Figure 1 and already introduced above, there are a number of different eye-movement parameters that can be used to address early and later memory retrieval processes during sentence reading (e.g. Inhoff and Radach, 1998; Rayner, 1998). When the eyes land on a word within a sentence during left-to-right reading, they remain relatively still for a particular amount of time, generally referred to by the term fixation duration. The first fixation on a word duration (FFD) is generally assumed to reflect early orthographic and lexical processing (Radach and Kennedy, 2004), but has also been shown to be sensitive for readily available predictive semantic (top-down) information for a given word (e.g. Roelke et al., 2020). The sum of all fixation durations before the eye leaves the word to the right is referred to by (first-pass) gaze duration (GD), which reflects later stages of word processing including lexical access. After leaving the word to the right, the eye may come back to the respective word and remain there for some time, which is further added into the total viewing duration (TVD). Such late eye movement measures reflect the time needed to provide full semantic integration of a word into the current language context (Radach and Kennedy, 2013).

Word length, frequency and word predictability from sentence context are generally accepted by the eye-tracking community to represent the most influential psycholinguistic variables on eye-movements (e.g. Engbert et al., 2005; Reichle et al., 2003). Word length is particularly affecting medium to late cognitive processes, while word frequency seems to affect all eye-movement measures (e.g. Kennedy et al., 2013). In psychology, word predictability from sentence context is typically estimated by cloze completion probabilities (Ehrlich and Rayner, 1981), which can be well approximated by language models (Shaoul et al., 2014). There are numerous studies examining the influence of predictability on eye movement measures, which found that predictability affects both early and late eye movement parameters: Therefore, Staub's (2015) review suggested that cloze-completion-probability-based predictability is an all-in variable confounding all sorts of predictive processes. We believe that language models provide the opportunity to understand different types of "predictability", therefore allowing for a deeper understanding of how experience shapes memory and how memory acts on retrieval than current models of eye-movement control (Reichle et al., 2003).

While Rapp (2014) proposed to use single-predictor regressions to approach corpus representativeness, a typical analytic approach to eye movements are multiple regression analyses. In this case, the fixation durations are approximated by a function:

$$f(x) = \Sigma \ \beta_n * x_n + \beta + error \quad (1)$$

In Formula 1, $x_n$ is the respective predictor variable such as length, frequency or predictability, and the free parameters are denoted by $\beta$. $\beta_n$ reflects the slope explained by the predictor variable n, while $\beta$ is the intercept of the regression equation. Error is minimized by ordinary least squares. In single-predictor analyses, correlation coefficients inform about the relative influence of a single variable. To see how much variance is explained by a single predictor, the correlation coefficient is often squared to give the amount of variance explained. Though the typical variance explained by a single predictor can vary as a function of the variables included in the regression model, an $r^2 = 0.0095$ ($r = 0.097$) for the frequency effect in GD is a good benchmark at this fixation-based level of analysis (e.g. Kliegl et al., 2006, Table 4).

A critical factor influencing multiple regression analyses is the correlation of the predictor variables itself. If they surpass an $r > 0.3$, multicollinearity starts to become problematic and the variance inflation factor reaches a first critical level of 1.09 (e.g. O'Brien, 2007). Therefore, we here relied on a sentence-perplexity- and word-probability-based stimulus selection procedure, to allow for an independent prediction of the major variables of interest, i.e. our word2vec-based word probabilities (WP) of the individual and the norm-based training corpora.

## 4 Methodology

### 4.1 Participants and corpora

For data protection purposes, we do not provide the exact age of the two German native participants, but they were 40-70 years old and male. Verbal IQ scores due to the IST-2000R were 106/115 for participant 1/2, respectively (Liepmann et al., 2007). Active vocabulary was estimated in the percentile ranks of 100/81 and passive vocabulary by 31/81 (Ibrahimović and Bulheller, 2005). The percentile ranks of reading fluency was 97/48 and comprehension percentiles of the participants were 52/31 (Schneider et al., 2007). Further assessment revealed a clearly differentiable interest profile peaking in medicine and nature as well as agriculture for Participant 1, vs. education and music for Participant 2 (Brickenkamp, 1990).

Individual reading behavior of both participants was recorded on a Microsoft surface tablet. During corpus collection, we also recorded eye movements by an eye tracker mounted on the tablet (60 Hz, EyeTribe Inc.). Therefore, future studies may constrain the individual corpora to only those text regions that have actually been looked at. Participants were instructed to spend a maximum of personal reading time on this tablet. They were instructed to examine content matching their personal interests over a period of two months. A java script collected screenshots, when the display changed. The screenshots were converted into greyscale images and rescaled by a multiplicator of 5. In addition, a median filter was used to remove noise, while contrast intensity was further enhanced. Finally, these pre-processed images were converted to ASCII by optical character recognition (Tesseract Software OCR; Smith, 2007). Next, we reviewed samples from the output data and the word-level confidence scores. At a confidence score of 80, a large majority of words (> 95%) were correctly identified by the OCR script, which we used as threshold for the inclusion into the corpora. In a final step, the data were cleaned from special characters and punctuations. The resulting corpus of participant 1 contained 486,721 tokens, and the corpus of participant 2 included 314,943 tokens. For computing individual word frequency (WF), we stemmed all words of the resulting token sample. To obtain comparable measures to norm-based WF, individual WF was calculated in per-million words and log10-transformed. The norm corpus was the German corpus of the Leipzig Wortschatz Project consisting of 1.1 billion tokens (Goldhahn et al., 2012).

### 4.2 Language models and stimulus selection

To generate stimulus materials containing words that are either predictable by the training corpora of one of the two participants or under the norm corpus, we trained word2vec models from the three different corpora, using genism 3.0.0 (Rehurek and Sojka, 2010)[1]. We trained skip-gram models with 100 hidden units in 10 iterations with a minimum frequency of 3 for the individual corpora and 5 for the norm-based corpus.

Stimulus selection started by computing WPs of 44,932 sentences of a German Wikipedia dump for the word2vec models under the three training corpora. For sentence selection, sentence perplexity (PP) scores were computed from the $WP_i$ for the n words of a sentence:

$$PP = 2^{-\frac{1}{n}\sum_{x_i}^{n} log_2 p(WP_i)} \quad (2)$$

PPs were rank-ordered for the three training corpora. To select sentences that are either predictable by one of the participants or by the norm corpus, we computed rank differences of these perplexity scores. Then we selected approximately one third of the sentences that provide a relatively low perplexity under one corpus, but a higher perplexity under the other two training corpora. Finally, we searched for words providing a WP = 0 under the individually trained language models and replaced them by highly probable words if this led to a meaningful and syntactically legal sentence. The 134 selected sentences contained 5-15 words ($M$ = 9.78, $SD$ = 1.92) and the 1,301 words ranged in length from 2 to 17 letters ($M$ = 5.49, $SD$ = 2.67). In the final stimulus set, there was a low correlation/multicollinearity of the individual and the norm-based WP (see Table 1 below), which allows to estimate the contribution of individual and the norm-based WP to word viewing durations independent from each other.

### 4.3 Eye tracking study

The eye tracking study was conducted approximately 1-2 months after the end of the individual corpus collection period. Eye tracking data were measured with a sampling frequency of 2000 Hz by an EyeLink 2k (SR Research Ltd.). The participant's head was positioned on a chin-rest and stimuli were presented in black color on a light-grey background (Courier New, size 16) on a 24-inch monitor (1680x1050 pixel). With a distance from eye to monitor of 67 cm, the size of a letter corresponded to 0.3° of visual angle. A three-point eye-position calibration was performed at the beginning of the experiment and after each comprehension question (see below). After an instruction screen, the 134 sentences were presented in randomized order in two blocks of 67 sentences, intermitted by a 5-minute break. For each sentence presentation, a fixation point appeared on the screen. Then, a sentence was presented in one line, with the first word located 0.5° of visual angle to the right. 2° to the right of the end of the sentence, the string "xxXxx" was presented. Participants were instructed to look at this string to indicate that sentence reading has been finished, which automatically initiated the continuation of the experiment. To make sure that participants read for comprehension, 17 yes/no and 13 open questions were presented after randomly selected sentences. All questions were answered correctly from both participants.

Right-eye fixation durations were analyzed by multiple regression analyses ($N$ = 1673). Fixation durations lower than 70 ms were excluded from analysis, as well as outliers longer than 800 ms for FFD, 1000 ms for GD, and 1500 ms for TVD. The first and last words of a sentence, as well as words with WPs or WFs of zero were excluded, leading to $N$ =1291 fixation events remaining for all analyses. The predictor variables in the multiple regressions were word length, norm-based and individual WF, as well as norm-based and individual WP. The word probabilities due to the two individual corpora, together with a stimulus and viewing time example is presented in Figure 2.

---
[1] https://radimrehurek.com/gensim/

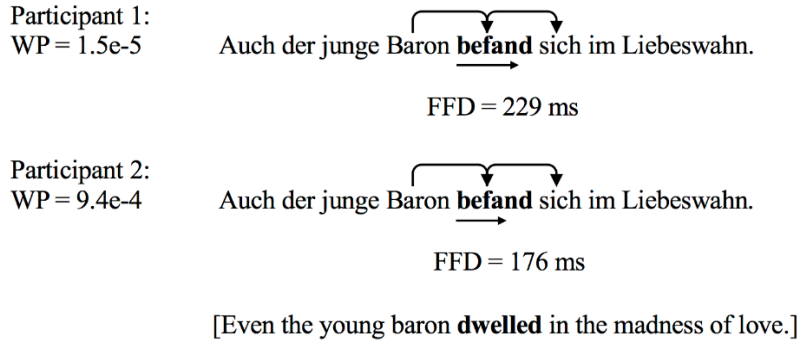

Figure 2. Individual word probabilities of an example word and the resulting viewing time: The language model trained by the corpus of Participant 1 provided a lower word probability (WP) than the corpus of Participant 2 in the example sentence. The higher word probability for Participant 2 predicts a faster first fixation duration (FFD).

## 5 Results

### 5.1 Correlation analysis and single-predictor regressions

The examination of the correlation between individual and norm-based WPs in Table 1 revealed that there was no significant correlation between these predictor variables. Therefore, our perplexity-based stimulus selection procedure will allow to examine whether these two predictors account for eye-movement variance independent from each other in the multiple regression analysis below. There were, however, typically large correlations between frequency and length and between the two frequency measures (e.g. Kliegl et al., 2006). Therefore, the question of whether length or frequency effects occur, cannot be answered unequivocally and e.g. frequency effects may be estimated by the predictor of word length in the multiple regression. All other correlation coefficients were smaller than 0.3, thus providing an uncritical level of multicollinearity.

|  | 1. | 2. | 3. | 4. | 5. | 6. | 7. | 8. |
|---|---|---|---|---|---|---|---|---|
| 1. Word length |  | -0.68 | -0.71 | 0.02 | -0.10 | 0.00 | **0.14** | **0.19** |
| 2. Norm-based WF | <.0001 |  | 0.80 | 0.09 | 0.13 | -0.05 | **-0.13** | **-0.16** |
| 3. Individual WF | <.0001 | <.0001 |  | 0.06 | 0.07 | -0.01 | **-0.10** | **-0.15** |
| 4. Norm-based WP | 0.4774 | 0.0013 | 0.0222 |  | **0.03** | -0.02 | -0.04 | -0.04 |
| 5. Individual WP | 0.0005 | <.0001 | 0.0174 | **0.3427** |  | *-0.12* | *-0.08* | *-0.06* |
| 6. FFD | 0.9447 | 0.0864 | 0.6284 | 0.4048 | *<.0001* |  | 0.74 | 0.37 |
| 7. GD | **<.0001** | **<.0001** | 0.0003 | 0.1277 | *0.0029* | <.0001 |  | 0.53 |
| 8. TVD | **<.0001** | **<.0001** | **<.0001** | 0.1610 | *0.0380* | <.0001 | <.0001 |  |

Table 1: Correlation of the predictor variables and the three word viewing time measures. Correlation coefficients are given above diagonal and correlation probability below.

When considering this correlation table as a single-predictor regression on the eye tracking data, there were effects of length, norm-based and individual word frequency in the GD and TVD data. While larger word length increased fixation durations, a larger norm-based and individual word frequency decreased the viewing times. No such effects were obtained in FFD data. Norm-corpus based WPs did not reveal any effects, while there were significant effects of individual WPs for FFD, GD, and TVD data, showing the largest correlation for FFD ($r = -0.12$), followed by GD ($r = -0.08$) and TVD ($r = -0.06$). Higher word probabilities decreased fixation durations.

## 5.2 Multiple regressions

The FFD analysis including all predictor variables provided a highly significant multiple regression model, $F(5,1285) = 4.629$, $p = 0.0003$ (Table 2). In all, it accounted for 1.77% of the variance. We obtained a significant effect of individual WP: Negative $t$-values indicated that high individual WP decreased FFD. Word frequency marginally failed to reach significance, with high frequency tending to diminish reading times.

|  | β | SE β | t | p |
|---|---|---|---|---|
| (Constant) | 268.793017 | 16.4469826 | 16.34 | <.0001 |
| Word length | -1.7378111 | 1.61136469 | -1.08 | 0.2810 |
| Norm-based WF | -5.2127476 | 2.83947993 | -1.84 | 0.0666 |
| Individual WF | 2.54936544 | 3.79117063 | 0.67 | 0.5014 |
| Norm-based WP | -73835.943 | 152204.436 | -0.49 | 0.6277 |
| Individual WP | -26123.236 | 6394.71736 | -4.09 | <.0001 |

Table 2: Results of the multiple regression analysis for FFD.

The multiple regression on GD data revealed a highly significant regression model, $F(5,1285) = 0.756$, $p < .0001$ (see Table 3), which in total accounted for 2.89% of the variance. We found a significant effect of word length, with longer words increasing GDs, as well as a significant effect of individual WP, with highly probable words reducing GDs.

|  | β | SE β | t | p |
|---|---|---|---|---|
| (Constant) | 245.517945 | 20.9540141 | 11.72 | <.0001 |
| Word length | 6.44767908 | 2.05293331 | 3.14 | 0.0017 |
| Norm-based WF | -4.988601 | 3.61759382 | -1.38 | 0.1681 |
| Individual WF | 4.76546508 | 4.83008007 | 0.99 | 0.3240 |
| Norm-based WP | -282605.29 | 193913.618 | -1.46 | 0.1453 |
| Individual WP | -18673.72 | 8147.08699 | -2.29 | 0.0221 |

Table 3: Results of the multiple regression analysis for GD.

For the TVD analysis, we obtained a highly significant multiple regression model, $F(5,1672) = 10.56$, $p < .0001$ (see Table 4). Overall, the predictors accounted for 3.95% of variance. Only word length provided a significant effect. Longer words lead to an increase of the TVD.

|  | β | SE β | t | p |
|---|---|---|---|---|
| (Constant) | 312.913912 | 36.015215 | 8.69 | <.0001 |
| Word length | 13.5909502 | 3.52852843 | 3.85 | 0.0001 |
| Norm-based WF | -6.498011 | 6.21782625 | -1.05 | 0.2962 |
| Individual WF | 1.23816155 | 8.30181612 | 0.15 | 0.8815 |
| Norm-based WP | -448879.23 | 333293.688 | -1.35 | 0.1783 |
| Individual WP | -17982.923 | 14003.0014 | -1.28 | 0.1993 |

Table 4: Results of the multiple regression analysis for TVD.

## 6 Discussion

To examine the representativeness of different training corpora for the learning experience, we here collected corpora for two individuals. These individual corpora were compared against a norm-based corpus (Goldhahn et al., 2012). We computed individual and norm-based semantic long-term memory structure based on word2vec models (Mikolov et al., 2013). Then we computed WPs for a sample of 45K sentences from Wikipedia and selected 134 sentences providing no significant correlation between the norm-based and the individual WPs to avoid multicollinearity in our multiple regression analyses on eye movement data.

Single-predictor and multiple regression analyses revealed that there are no significant effects of norm-based WP with any eye movement measure. One possible reason for these zero findings could be that we replaced words with a WP = 0 by words that are highly expectable under the individual, but not under the norm-based corpus. This might have increased the sensitivity for successful predictions of individual corpora. The stimulus selection procedure optimized the eye-tracking experiment for the representativeness for our participants' individual knowledge structure. This may compromise the representativeness for other types of knowledge.

The individual WP, in contrast, revealed reliable single-predictor effects. These effects were largest in FFD data and decreased for later GD and TVD data. When comparing the 1.44% of explained variance of individual WP in the FFD data to the total variance of 1.77% explained in the multiple regression, this result pattern suggests that most of the variance was explained by individual WP. The slight increase in explained variance primarily results from norm-based WF, which marginally failed to reach significance in the multiple regression analysis on FFD data. In general, the amount of explained variance in these analyses are comparable to other multiple regression studies predicting each fixation duration, without aggregating across eye-movement data (e.g. Kliegl et al., 2006).

Single and multiple regression analyses revealed that the effect of individual WP tends to become smaller, but apparent in GD data. This suggests that individual corpora are most suitable to predict early to mid-latency eye movement measures (cf. Figure 1). In the multiple regression analysis of GD data, an additional effect of word length was observed, confirming the finding that length has a larger effect on such later eye movement parameters (e.g. Kennedy et al., 2013), because multiple fixations are more likely in longer words. In TVD data, we observed no effect of individual WP, but a large effect of word length in the multiple regression analyses.

The largest limitation of the present study is the low statistical power of eye tracking data from two participants only. As Rapp (2014) already noted, the collection of individual corpora is effortful, but we think that this effort was worthwhile, even when the present study relied on a limited amount of statistical power. We were positively surprised by the reliable effect of individual WP in early and mid-range eye movement parameters. Nevertheless, the present work should be seen as a pilot study that will hopefully encourage further examinations of individual corpora. But there is also a second power issue that makes these results convincing. Our norm-based training corpus was at least 140 times larger than the individual training corpora. Therefore, we think that this is sound evidence that representativeness of a corpus for individual long-term memory structure can outperform size in predicting individual reading performance (e.g. Banko and Brill, 2001, inter alia).

One reason for our conclusion that individual corpora may better predict eye movements lies in the time period, in which the text corpus reflecting human experience was acquired. The individual corpora were collected in a two-month time period that preceded the eye tracking study by about 1-2 months. Therefore, the individual corpora may primarily reflect more recently acquired knowledge. Ericsson and Kintsch (1995) proposed multiple buffer stores in their theory of long-term working memory (cf. Kintsch and Mangalath, 2011). Recently acquired knowledge is held in these buffers before it is integrated into long-term memory. Therefore, participants can stop reading a text, and when carrying on reading later, only the first sentences are read slower as compared to continuous reading of these texts (Ericsson and Kintsch, 1995). Our early eye-movement effects may be explained by the proposal that the recently acquired knowledge still resides in Ericsson and Kintsch's (1995) long-term buffer stores. While they argue that football knowledge can well predict comprehension of football-related texts, for instance, it is hard to answer the question of which knowledge has been acquired in which time period. First, individual corpora may help to free such studies from the investigation of one particular type of knowledge, because each individual corpus reflects the knowledge of this individual. Second, individual corpora collected at different time periods may provide a novel approach to the question of how long information may persist in these knowledge buffers. Our results suggest that information still residing in long-term memory buffers elicits faster and more efficient memory retrieval.

There are several studies, in which participants are required to write diaries, which can be considered as extremely small individual corpora (e.g. Campbell & Pennebaker, 2003). Another example is the task to write emails to predict individual traits: For instance, Oberlander and Gill (2006) found that participants with high extraversion tend to use "will not" for expressing negation, while participants with low extraversion tend to use "not really" (see Johannßen and Biemann, 2018, for a recent overview). While such studies focused on the language output, the present study provided two input variables. First, individual corpora allow to estimate individual experience. Second, we selected the materials based on a language model to specifically capture the individual experience of the participants in the eye-tracking experiment. The language models can be considered an algorithmic approach to the neurocognitive system between the inputs and the output.

Much as differences of the cognitive architecture of the participants, there are "interindividual" differences between language models. For instance, n-gram models may reflect the capability of participants to remember particular words in context of specific other words, while neural network models more closely reflect the human capability to generalize from experiences (e.g. Hofmann et al., 2020; McClelland & Rogers, 2003). Thus, comparing such models with respect to the question of which model predicts which participant may provide information about the cognitive architecture of the respective participant. With respect to human intelligence testing, individual corpora should be a suitable approach to face Catell's (1943, p. 157) challenge of "freeing adult tests from assumptions of uniform knowledge".

For future work, we would like to proceed in three directions. First, we would like to improve the collection procedure: the corpora collected via screenshots and OCR contain a high number of artifacts stemming from non-textual material, as well as non-contiguous texts as a result from complex webpage layouts. Second, we like to increase the number of participants in future studies. Third, it would be interesting to compare our word2vec results with more recent contextual embeddings such as BERT (Devlin et al., 2019), which have been shown to achieve better performance across a wide range of natural language processing tasks than language models with static word embeddings. While it is non-trivial to use BERT's bi-directional architecture and its masking mechanism for language modelling tasks, Salazar et al., (2020) have recently shown how to obtain prediction values for BERT and other architectures trained with masking loss. Subword representations as used in BERT may also help to compensate OCR-based errors, when only a few letters have been falsely recognized. On the downside, it is questionable whether the present corpus sizes of 300/500K token are large enough to obtain reliable estimates for the large number of BERT's parameters. A potential solution is to rely on a BERT model pre-trained by a large corpus, and to use the individual corpora to fine-tune the language model. Though such fine tuning may enhance the predictions over the pre-trained model only, such an approach would mix norm-based and individual corpus information. The aim of the present study, in contrast, was to focus on the comparison of norm-based vs. strictly individual corpora, so we leave this extension for future work[2].

# References

Banko, M., & Brill, E. (2001). Scaling to very very large corpora for natural language disambiguation. In *Proceedings of the 39th annual meeting on association for computational linguistics* (pp. 26–33).

Bhatia, S. (2017). Associative judgment and vector space semantics. *Psychological Review*, *124*(1), 1–20.

Blei, D. M., Ng, A. Y., & Jordan, M. I. (2003). Latent dirichlet allocation. *Journal of Machine Learning Research*, *3*, 993–1022.

Brickenkamp, R. (1990). *Generelle Interessenskala*. Göttingen: Hogrefe.

Brysbaert, M., & New, B. (2009). Moving beyond Kučera and Francis: A critical evaluation of current word frequency norms and the introduction of a new and improved word frequency measure for American English. *Behavior Research Methods*, *41*(4), 977–990.

Campbell, R. S., & Pennebaker, J. W. (2003). The secret life of pronouns: Flexibility in writing style and physical health. *Psychological Science*, *14*(1), 60–65.

Catell, R. B. (1943). The measurement of adult intelligence. *Psychological Bulletin*, *40*(3), 153–193.

Deerwester, S., Dumais, S. T., Furnas, G. W., Landauer, T. K., & Harshman, R. (1990). Indexing by latent semantic analysis. *Journal of the American Society for Information Science*, *41*(6), 391–407.

---

[2] This research was partially supported by a grant of the Deutsche Forschungsgemeinschaft to MJH (DFG-Gz. 5139/2-2). We like to thank Saskia Pasche, Steffen Remus, Dirk Johanßen, and Christian Vorstius for help during data collection and analyses.

Demberg, V., & Keller, F. (2008). Data from eye-tracking corpora as evidence for theories of syntactic processing complexity. *Cognition*, *109*(2), 193–210.

Devlin, J., Chang, M.-W., Lee, K., Toutanova, K. (2019): BERT: Pre-training of deep bidirectional transformers for language understanding. Proceedings of the 2019 Conference of the North American Chapter of the Association for Computational Linguistics: Human Language Technologies, Volume 1 (Long and Short Papers) (pp. 4171–4186).

Ehrlich, S. F., & Rayner, K. (1981). Contextual effects on word perception and eye movements during reading. *Journal of Verbal Learning and Verbal Behavior*, *20*(6), 641–655.

Engbert, R., Nuthmann, A., Richter, E. M., & Kliegl, R. (2005). SWIFT: a dynamical model of saccade generation during reading. *Psychological Review*, *112*(4), 777–813.

Ericsson, K. A., & Kintsch, W. (1995). Long-term working memory. *Psychological Review*, *102*(2), 211–245.

Frank, S. (2009). Surprisal-based comparison between a symbolic and a connectionist model of sentence processing. In *Proceedings of the annual meeting of the Cognitive Science Society* (pp. 1139–1144).

Frank, S. L., & Bod, R. (2011). Insensitivity of the human sentence-processing system to hierarchical structure. *Psychological Science*, *22*(6), 829–834.

Goldhahn, D., Eckart, T., & Quasthoff, U. (2012). Building large monolingual dictionaries at the leipzig corpora collection: From 100 to 200 languages. In *Proceedings of the 8th International Conference on Language Resources and Evaluation* (pp. 759–765).

Griffiths, T. L., Steyvers, M., & Tenenbaum, J. B. (2007). Topics in semantic representation. *Psychological Review, 114*(2), 211–244.

Hofmann, M. J., Biemann, C., & Remus, S. (2017). Benchmarking n-grams, topic models and recurrent neural networks by cloze completions, EEGs and eye movements. In B. Sharp, F. Sedes, & W. Lubaszewsk (Eds.), *Cognitive Approach to Natural Language Processing* (pp. 197–215). London, UK: ISTE Press Ltd, Elsevier.

Hofmann, M. J., Biemann, C., Westbury, C. F., Murusidze, M., Conrad, M., & Jacobs, A. M. (2018). Simple co-occurrence statistics reproducibly predict association ratings. *Cognitive Science*, *42*, 2287–2312.

Hofmann, M. J., Kuchinke, L., Biemann, C., Tamm, S., & Jacobs, A. M. (2011). Remembering words in context as predicted by an associative read-out model. *Frontiers in Psychology*, *2*(252), 1–11.

Hofmann, M. J., Remus, S., Biemann, C., & Radach, R. (2020). Language models explain word reading times better than empirical predictability. Retrieved from https://psyarxiv.com/u43p7/download?format=pdf

Ibrahimovic, N., & Bulheller, S. (2005). *Wortschatztest – aktiv und passiv*. Frankfurt am Main: Harcourt.

Inhoff, A. W., & Radach, R. (1998). Definition and computation of oculomotor measures in the study of cognitive processes. In G. Underwood (Ed.), *Eye Guidance in Reading and Scene Perception* (pp. 29–53). Oxford, England: Elsevier Science.

Johannßen, D., & Biemann, C. (2018). Between the lines: Machine learning for prediction of psychological traits- A survey. In *International Cross-Domain Conference for Machine Learning and Knowledge Extraction* (pp. 192-211). Springer, Cham.

Johns, B. T., Jones, M. N., & Mewhort, D. J. K. (2019). Using experiential optimization to build lexical representations. *Psychonomic Bulletin & Review*, *26*(1), 103-126.

Jacobs, A. M. (2019). Sentiment analysis for words and fiction characters from the perspective of computational (neuro-)poetics. *Frontiers in Robotics and AI*, *6*(53), 1–13.

Kennedy, A., Pynte, J., Murray, W. S., & Paul, S. A. (2013). Frequency and predictability effects in the Dundee Corpus: An eye movement analysis. *Quarterly Journal of Experimental Psychology*, *66*(3), 601–618.

Kintsch, W., & Mangalath, P. (2011). The construction of meaning. *Topics in Cognitive Science*, *3*(2), 346–370.

Kliegl, R., Nuthmann, A., & Engbert, R. (2006). Tracking the mind during reading: the influence of past, present, and future words on fixation durations. *Journal of Experimental Psychology. General*, *135*(1), 12–35.

Liepmann, D., Beauducel, A., Brocke, B., & Amthauer, R. (2007). *Intelligenz-Struktur-Test 2000 R (I-S-T 2000 R)*. Göttingen: Hogrefe.

Mandera, P., Keuleers, E., & Brysbaert, M. (2017). Explaining human performance in psycholinguistic tasks with models of semantic similarity based on prediction and counting: A review and empirical validation. *Journal of Memory and Language*, *92*, 57–78.

McClelland, J. L., & Rogers, T. T. (2003). The parallel distributed processing approach to semantic cognition. *Nature Reviews Neuroscience*, *4*(4), 310–322.

McDonald, S. A., & Shillcock, R. C. (2003). Eye movements reveal the on-line computation of lexical probabilities during reading. *Psychological Science*, *14*(6), 648–652.

Mikolov T. (2014). *Statistical language models based on neural networks* (PhD Thesis). Brno University of Technology, Brno.

Mikolov, T., Chen, K., Corrado, G., & Dean, J. (2013). Efficient estimation of word representations in vector space. In *Proceedings of Workshop at ICLR* (pp. 1–12).

Oberlander, J., & Gill, A. J. (2006). Language with character: A stratified corpus comparison of individual differences in e-mail communication. *Discourse Processes*, *42*(3), 239–270.

O'Brien, R. M. (2007). A caution regarding rules of thumb for variance inflation factors. *Quality and Quantity*, *41*(5), 673–690.

Paller, K. A., & Wagner, A. D. (2002). Observing the transformation of experience into memory. *Trends in Cognitive Sciences*, *6*(2), 93–102.

Pynte, J., New, B., & Kennedy, A. (2008). On-line contextual influences during reading normal text: A multiple-regression analysis. *Vision Research*, *48*(21), 2172–2183.

Radach, R., Inhoff, A. W., Glover, L., & Vorstius, C. (2013). Contextual constraint and N + 2 preview effects in reading. *Quarterly Journal of Experimental Psychology*, *66*, 619–633.

Radach, R., & Kennedy, A. (2004). Theoretical perspectives on eye movements in reading: Past controversies, current issues, and an agenda for future research. *European Journal of Cognitive Psychology, 16*(1–2), 3–26.

Radach, R., & Kennedy, A. (2013). Eye movements in reading: Some theoretical context. *Quarterly Journal of Experimental Psychology*, *66*(3), 429–452.

Rapp, R. (2014). Using collections of human language intuitions to measure corpus representativeness. *COLING 2014 - 25th International Conference on Computational Linguistics, Proceedings of COLING 2014: Technical Papers*, 2117–2128.

Rayner, K. (1998). Eye movements in reading and information processing: 20 years of research. *Psychological Bulletin*, 124(3), 372–422.

Rehurek, R., & Sojka, P. (2010). Software framework for topic modelling with large corpora. In R. Witte, H. Cunningham, J. Patrick, E. Beisswanger, E. Buyko, U. Hahn, K. Verspoor, & A. R. Coden (Eds.)*, Proceedings of the LREC 2010 Workshop on New Challenges for NLP Frameworks* (pp. 45–50). Valletta, Malta: ELRA.

Reichle, E. D., Rayner, K., & Pollatsek, A. (2003). The E-Z reader model of eye-movement control in reading: comparisons to other models. *The Behavioral and Brain Sciences*, *26*(4), 445–476.

Reilly, R. G., & Radach, R. (2006). Some empirical tests of an interactive activation model of eye movement control in reading. *Cognitive Systems Research*, *7*(1), 34–55.

Roelke, A., Vorstius, C., Radach, R., & Hofmann, M. J. (2020). Fixation-related NIRS indexes retinotopic occipital processing of parafoveal preview during natural reading. *NeuroImage*, *215*, 116823.

Salazar, J., Liang, D., Nguyen, T. Q., & Kirchhoff, K. (2019). Masked Language Model Scoring. Retrieved from https://arxiv.org/pdf/1910.14659.pdf

Shaoul, C., Baayen, R. H., & Westbury, C. F. (2015). N-gram probability effects in a cloze task. *The Mental Lexicon*, *9*(3), 437–472.

Schneider, W., Schlagmüller, M., & Ennemoser, M. (2007). LGVT 6-12: Lesegeschwindigkeits- und -verständnistest für die Klassen 6-12 (p. 6). Göttingen: Hogrefe.

Smith, R. (2007). An overview of the Tesseract OCR Engine. In *Proceedings of the Ninth International Conference on Document Analysis and Recognition (ICDAR 2007),* Vol. 2. IEEE Computer Society, Washington, DC, USA, 629–633.

Smith, N. J., & Levy, R. (2013). The effect of word predictability on reading time is logarithmic. *Cognition*, *128*(3), 302–319.

Staub, A. (2015). The effect of lexical predictability on eye movements in reading: Critical review and theoretical interpretation. *Language and Linguistics Compass*, *9*(8), 311–327.